\documentclass[]{style}

\usepackage[toc,page,header]{appendix}
\usepackage{amsmath}
\usepackage{amssymb}
\usepackage{float}

\title{Repo2Skill-Evo: Repository Skills Go Stale in Silence}

\affiliation[1]{ByteDance}
\affiliation[2]{Peking University}
\affiliation[3]{Beijing Jiaotong University}

\contribution{Full author list in Contributions}

\abstract{
Large language model (LLM) agents increasingly operate over evolving software repositories, where success depends on repository-specific procedural knowledge: which APIs to call, which scripts to run, and which conventions the current release expects. Agent skills externalize this knowledge into reusable units, and prior work shows that they can improve agent performance. What remains unclear is whether that improvement is durable. The same version specificity that makes a skill useful also makes it fragile: after a release, it may become stale without raising any explicit signal, while continuing to provide obsolete guidance. Externalizing knowledge into a skill can therefore make its decay invisible.

We study whether agents can keep this externalized knowledge current. Repo2Skill-Evo casts each release transition as a skill-maintenance task: given a $V_1$ skill set and the official $V_1\!\rightarrow\!V_2$ patch, an agent must update obsolete skill content while preserving guidance that remains valid. Across 57 real-world repositories and 105 selected release transitions, every evaluated transition invalidates part of the $V_1$ skill set. Yet six frontier agents reach only 29.9\%--69.7\% avg@3 macro $F_1$ under a patch-grounded removal metric that balances stale-content recall against over-editing precision. Across runs, two opposing errors dominate: incomplete coverage of affected files in the skill set leaves stale content untouched, while overbroad editing is associated with higher recall but lower precision.
\textbf{Repository skills go stale in silence, and even frontier agents cannot reliably maintain them.}
}

\begin{document}
\maketitle

\section{Introduction}
\label{sec:intro}

Large language model~(LLM) agents increasingly act through tools and interactive environments~\citep{react,toolformer,agentbench}, and repository-level software engineering has become one of their most demanding settings. To complete realistic repository tasks, an agent must inspect code, read documentation, edit files, and run scripts in an evolving codebase~\citep{swebench,sweagent}. Success therefore depends not only on the underlying model, but also on repository-specific procedural knowledge: which APIs to call, which scripts to run, how modules are configured, and which conventions the current release expects. Extracting this knowledge from scratch for every task is costly and unreliable, since it requires cross-file retrieval and structural understanding that even strong agents do not consistently achieve~\citep{repobench,repomirage}.

Agent skills offer a compact alternative. They externalize reusable procedural knowledge into on-demand units that agents can retrieve and follow~\citep{sok_skills,skillsbench,swe_skills_bench}. Prior work shows that skills can improve downstream agent performance~\citep{skillsbench,swe_skills_bench,skillgenbench}, and repository-local skills are beginning to appear beyond benchmark settings~\citep{agent_skills_overview,pytorch_skills}. Skills are thus becoming persistent agent-facing resources across tasks and releases.

\begin{figure}[!t]
\centering
\includegraphics[width=\textwidth]{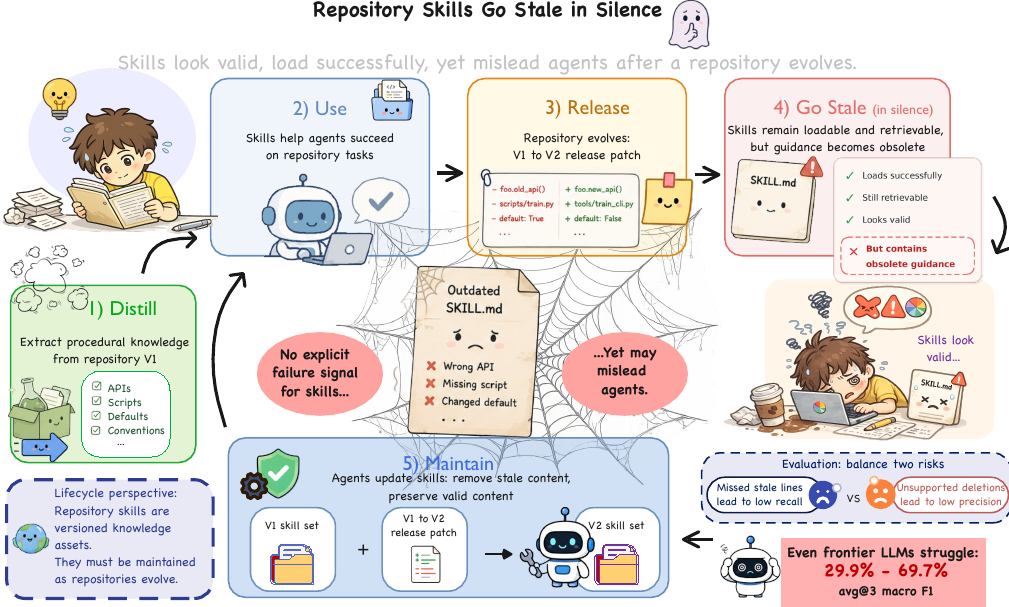}
\caption{Lifecycle of repository skills and the Repo2Skill-Evo maintenance setting. Repo2Skill distills repository-specific procedural knowledge from a $V_1$ repository into a traceable skill set. After the repository evolves from $V_1$ to $V_2$, previously valid skill content may become silently stale: it remains loadable and retrievable while continuing to provide guidance that no longer matches the current repository state. Given the fixed $V_1$ skill set and the official release patch, a maintenance agent produces the updated skill set $\mathcal{S}_{V_2}$. Repo2Skill-Evo evaluates whether the agent removes stale content while preserving guidance that remains valid.}
\label{fig:overview}

\end{figure}

We take this utility as our starting point and ask: once a repository evolves, can agents keep its skills from silently going stale? A repository skill is useful precisely because it encodes version-specific knowledge, but that same specificity ties its correctness to a particular repository state. After a release, a skill distilled from the previous version may remain loadable and retrievable while continuing to provide guidance that no longer matches the current repository state. Because no explicit signal marks this mismatch, the outdated guidance may continue to be retrieved and followed. We refer to this release-driven failure mode as~\emph{silent staleness}. Externalizing version-specific knowledge into a skill preserves its form while rendering its decay invisible. Figure~\ref{fig:overview} summarizes this lifecycle and the resulting maintenance setting.

We therefore treat repository-grounded skills as versioned knowledge assets and center our study on their maintenance. Although related to code migration, the maintained object is the agent-facing guidance derived from a repository, not the repository code itself. Because a release may invalidate only part of a skill package, maintenance requires removing or revising stale content while preserving what remains valid.

To study this problem at scale, Repo2Skill-Evo combines three components. First, Repo2Skill uses staged distillation and expert refinement to produce a fixed $V_1$ skill set. Second, given this skill set and the official $V_1\!\rightarrow\!V_2$ release patch, a maintenance agent must remove or revise obsolete skill content while preserving guidance that remains valid. Third, a strictly patch-grounded headline metric rewards the removal or revision of patch-verified obsolete $V_1$ skill lines while penalizing edits to other $V_1$ skill lines, complemented by a rubric-based NL Judge that provides a five-dimensional assessment of the resulting $V_2$ skill set~\citep{geval,llm_judge,prometheus}.

A focused utility study further motivates the maintenance problem. Across ten repositories randomly sampled from the corpus, skills yield the largest absolute gains on tasks with the lowest baseline utility. Within this sample, the pattern suggests that maintenance may matter most where agents have limited repository knowledge without external support.

Our main evaluation across 57 repositories and 105 selected release transitions yields two findings. \textbf{First, agents do not reliably maintain repository skills across releases.} Each selected transition includes a $V_1$ skill set with patch-verified stale content (median: 92 lines), yet six frontier agents achieve only 29.9\%--69.7\% avg@3 macro $F_1$. \textbf{Second, failures reflect complementary localization and edit-selection bottlenecks.} Incomplete coverage of affected skill files leaves stale content untouched, whereas broader edits are associated with higher recall but lower patch-grounded precision. Oracle skill-file localization on the 20 hardest transitions improves performance, but substantial residual errors remain, indicating that localization alone does not resolve within-file evidence tracing and edit selection.

This paper makes three contributions:
\begin{itemize}

\item We characterize \textbf{silent staleness}, a release-driven temporal failure mode of repository-grounded skills, and frame these skills as versioned knowledge assets.

\item We introduce \textbf{Repo2Skill-Evo}, a patch-grounded maintenance benchmark covering 57 repositories and 105 official release transitions, with fixed $V_1$ skill sets and patch-verified obsolete-line annotations.

\item We evaluate six frontier agents and identify two key maintenance bottlenecks: incomplete localization of affected skill files and imperfect edit selection.

\end{itemize}

\section{Related Work}
\label{sec:related}

\subsection{Repository-Level Agents and Software Evolution}

Repository-level software engineering is a central setting for evaluating LLM agents. SWE-bench casts real GitHub issues as end-to-end repair tasks that require understanding a codebase and producing test-passing patches~\citep{swebench}, while SWE-agent shows that the agent--computer interface shapes how agents navigate, edit, and test repositories~\citep{sweagent}. Other systems study issue resolution, localization, and repair scaffolding~\citep{autocoderover,agentless}. RepoBench emphasizes cross-file context~\citep{repobench}, and RepoMirage shows that broad repository access does not necessarily yield structural understanding~\citep{repomirage}.

A parallel line studies software evolution directly, including library migration, API refactoring, and migration detection~\citep{java_library_migration,python_library_migration,api_refactoring,migrationminer}. Recent agent benchmarks extend from static repair to release-note-derived evolution tasks~\citep{swe_evo}, continuous-integration loops~\citep{sweci}, and chained release-level upgrades~\citep{swechain}. These works center on source code or downstream client usage. Repo2Skill-Evo instead maintains the agent-facing knowledge that describes how to use a repository. Because this knowledge is a separate artifact, its maintenance does not reduce to code migration.

\subsection{Agent Skill Construction and Utility}

Agent skills package reusable procedural knowledge into modular units that agents retrieve when relevant. Surveys organize this area around the skill lifecycle, spanning representation, acquisition, retrieval, and evolution~\citep{sok_skills,agent_skills_survey}. Skills can be generated from source corpora~\citep{skillgenbench} and improved through execution and verifier feedback~\citep{evoskills,skillrevise}. Utility benchmarks report heterogeneous effects: curated skills can improve performance, whereas self-generated or mismatched skills do not reliably help~\citep{skillsbench,swe_skills_bench,skillgenbench}.

Most of this work evaluates skill construction or use within a fixed environment snapshot. Repository-local skills also appear outside benchmarks. \texttt{SKILL.md} anchors an emerging cross-agent format~\citep{agent_skills_overview}, and PyTorch ships skills for dispatch-macro migration, Metal kernel implementation, AOTInductor debugging, and docstring maintenance~\citep{pytorch_skills}. These examples establish repository skills as practical agent-facing artifacts, but do not establish that they remain valid after their repositories advance to new releases.

\subsection{Agent Skill Lifecycle and Maintenance}

Software engineering has long recognized that natural-language artifacts can drift out of sync with code, as in code-comment inconsistency~\citep{code_comment_inconsistency}. A repository skill serves as an agent-facing analogue carrying direct operational consequences. It is not merely reviewed by human maintainers but can also be retrieved and executed by autonomous agents. Runtime feedback can support skill repair after an observed execution failure~\citep{contractskill}, but silent staleness may provide no error log to react to.

Recent work addresses distinct parts of the skill lifecycle. SkillGuard detects environment-contract violations and uses them to localize repair~\citep{skilldrift}. SkillOps and Library Drift tackle concerns surrounding skill libraries. SkillOps mitigates technical debt via typed contracts and graph-structured maintenance~\citep{skillops}, while Library Drift investigates retrieval degradation stemming from unregulated expansion within self-evolving skill banks~\citep{library_drift}. SkillHone focuses on iterative revision by preserving decision histories and evaluation evidence across sessions~\citep{skillhone}. An empirical study complements these systems by characterizing how registry and personal-use skills are reused, adapted, and maintained in practice~\citep{skill_maintenance_empirical}.

Repo2Skill-Evo isolates release-conditioned maintenance. The release change is provided as an official patch, and the challenge is to map its implications onto a fixed repository-grounded skill set while preserving unaffected guidance. Unlike skill internalization~\citep{skillinternalize}, the maintained object remains an external artifact whose validity is tied to a software release.

\section{Repo2Skill-Evo: Maintaining Repository Skills across Releases}

\label{sec:repo2skill}

\begin{figure}[!b]
\centering
\includegraphics[width=\textwidth]{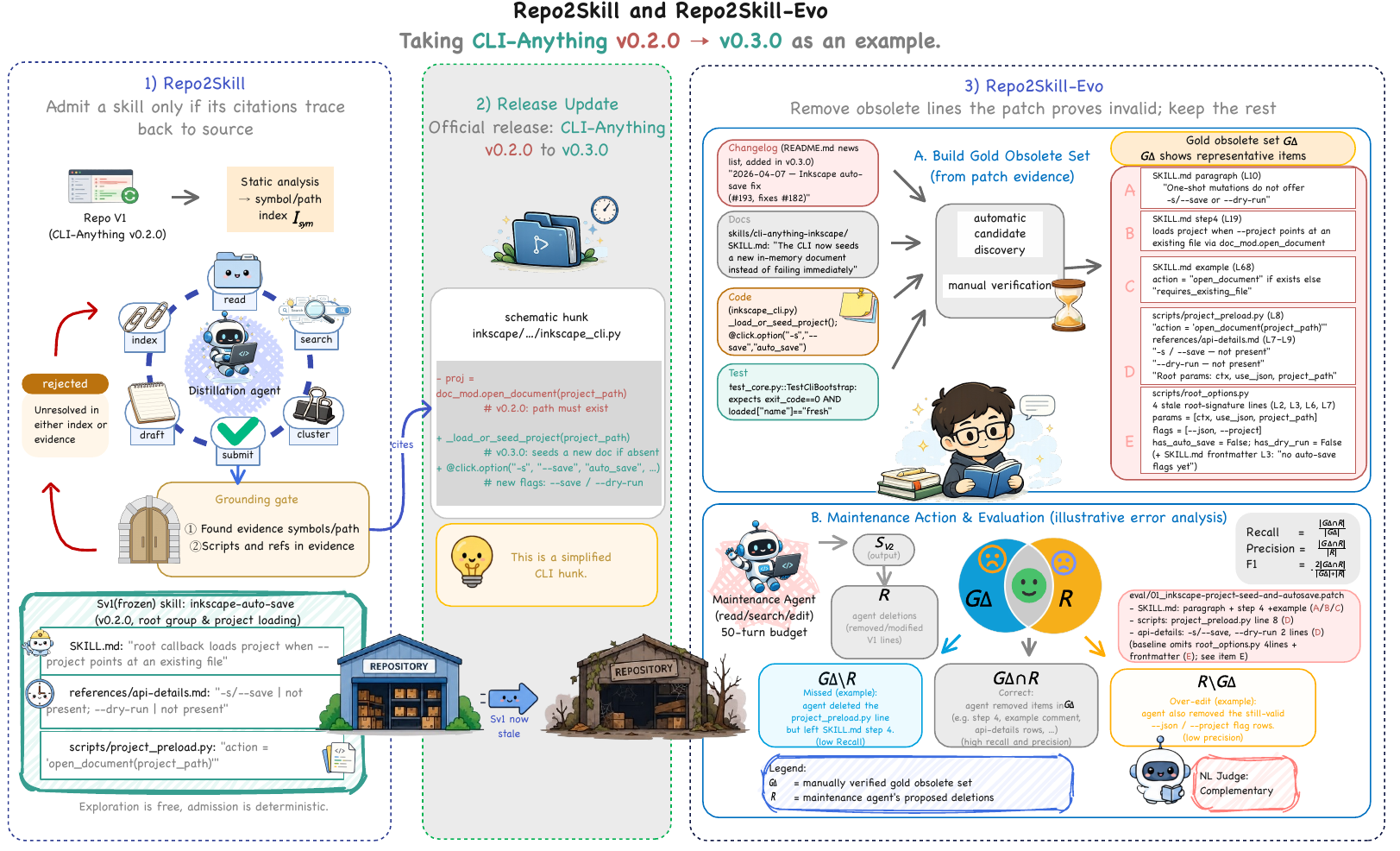}
\caption{Repo2Skill and Repo2Skill-Evo on a release example. Using a simplified \texttt{CLI-Anything} $v0.2.0\!\rightarrow\!v0.3.0$ update, the figure illustrates three stages: (1) Repo2Skill generates candidate $V_1$ skill packages, after which an expert selects, revises, and verifies one to produce the fixed $V_1$ skill set; (2) the release patch changes a documented schema behavior; and (3) Repo2Skill-Evo asks a maintenance agent to remove obsolete lines while preserving valid content. Evaluation uses the patch-grounded gold obsolete set $G_{\Delta}$ (Section~\ref{sec:metric}). The same construction is applied across all 105 release transitions.}
\label{fig:pipeline}

\end{figure}

Repo2Skill-Evo has two components, a procedure for obtaining a fixed, traceable $V_1$ skill set to maintain, and a release-level task that asks whether agents can keep that set current. We describe the skill set first, then the maintenance task, scaffold, and evaluation. Figure~\ref{fig:pipeline} summarizes the full pipeline.

\subsection{Repo2Skill: Distilling a Traceable Skill Set}
\label{sec:distill}

Before asking whether agents can keep a skill set current, we first need a fixed $V_1$ skill set whose repository-specific claims can be traced to source evidence. Repo2Skill serves this role by distilling repository-specific procedural knowledge into grounded skill packages (Figure~\ref{fig:pipeline}, left), providing the common starting point for the maintenance experiments that follow.

\paragraph{Skill representation.} A skill is a self-contained package that documents a reusable procedure in \texttt{SKILL.md}, with optional runnable \texttt{scripts} and supporting \texttt{references}. Repo2Skill adapts this format to repository settings by distilling repository-specific procedures into a skill set $\mathcal{S}=\{s_1,\dots,s_m\}$. Provenance is recorded during distillation. The agent logs the source paths and symbols behind each claim as external metadata, which supports the grounding gate and later tracing of content affected by repository changes.

\paragraph{The distillation agent.}
\label{sec:orchestration}

The agent's \textsc{analyze} stage performs static analysis to build a symbol index $I_{\mathrm{sym}}$ that maps repository symbols and file paths to source locations. This index anchors subsequent logic analysis and material collection. The distillation agent then runs a think--act--observe loop over a workspace state
\[
w_t=(\mathcal{H}_t,A_t,L_t,M_t,\mathcal{S}_t),
\]
where $\mathcal{H}_t$ is the interaction history and $A_t$, $L_t$, $M_t$, and $\mathcal{S}_t$ are the accumulated analysis report, logic read-map, material documents, and skill package. At each step, the agent chooses from
\[
\mathcal{A}_{\mathrm{distill}}=
\{\textsc{analyze},\textsc{logic},\textsc{materials},
\textsc{draft},\textsc{validate},\textsc{finish}\}.
\]

\textsc{Analyze} runs first to construct $I_{\mathrm{sym}}$, after which \textsc{logic} derives a routing read-map and \textsc{materials} collects repository-grounded evidence over the same repository partitions. \textsc{Draft} jointly uses these intermediate products to author the skill package, \textsc{validate} applies the grounding gate, and \textsc{finish} terminates the run once validation succeeds. The stages follow this order by default, with earlier stages revisited only for recovery after a failure or validation rejection. Figure~\ref{fig:pipeline} illustrates this process through a simplified worked example.

\paragraph{Grounding gate.}
\label{sec:grounding}

After \textsc{draft}, \textsc{validate} applies structural and grounding checks to the candidate skill set by verifying the source paths and symbols referenced in the generated \texttt{SKILL.md} files and \texttt{scripts} against the $V_1$ repository. For candidates meeting the minimum reference count, validation fails once the fraction of unverifiable references exceeds a preset threshold, and only the candidates that pass proceed to expert selection and refinement. The gate therefore establishes source traceability, and its scope stops at reference resolution, since a referenced path or symbol can resolve even when the surrounding guidance uses it incorrectly. Semantic correctness and procedural coverage accordingly fall to the expert, who reviews both during selection and refinement.

\paragraph{Candidate generation and selection.} For each transition, Claude-opus-4.6~\citep{claude_opus46} and GPT-5.4~\citep{gpt54} each produce one candidate $V_1$ skill set under the same staged Repo2Skill procedure, using only the pre-release repository. An expert selects the stronger candidate, then revises and verifies it against the $V_1$ repository. The resulting ${\mathcal{S}_{V_1}}$ is frozen before gold-label construction begins and is shared by all maintenance agents. Appendix~\ref{app:v1-skills} details this process.

\subsection{Maintenance Task}
\label{sec:evo-task}

Once a repository advances from $V_1$ to $V_2$, parts of its $V_1$ skill set may no longer match the updated interfaces, defaults, file layouts, or procedures. Maintenance therefore requires mapping patch-level changes to the affected guidance and making targeted updates while preserving content that remains valid. Repo2Skill-Evo operationalizes this problem over official release transitions. The right panel of Figure~\ref{fig:pipeline} summarizes the maintenance setting and evaluation.

\paragraph{Task definition.} Each maintenance instance is a tuple $(\mathcal{S}_{V_1}, \Delta_{V_1\to V_2}, \mathcal{U}_{\mathrm{maint}})$ formed by the fixed $V_1$ skill set $\mathcal{S}_{V_1}$, the official release patch $\Delta_{V_1\to V_2}$, and the maintenance tool interface $\mathcal{U}_{\mathrm{maint}}$ available to the agent. The agent produces an updated skill set $\mathcal{S}_{V_2}$ and is evaluated on the set-level transformation $\mathcal{S}_{V_1}\rightarrow\mathcal{S}_{V_2}$. Successful maintenance requires the three capabilities below.

\begin{itemize}

\item \textbf{Patch-to-skill localization.} Identify the skill content affected by patch-level changes to APIs, files, commands, configurations, or usage procedures.

\item \textbf{Obsolete-knowledge removal.} Remove or revise stale references, signatures, examples, tables, and prose that are no longer valid under $V_2$.

\item \textbf{Information retention.} Preserve skill content that remains valid, avoiding unsupported rewrites, dropped context, or wholesale regeneration that discards useful repository-specific knowledge.

\end{itemize}

These requirements motivate our patch-grounded removal metric and complementary final-state NL Judge.

\subsection{Agent Scaffold}

To attribute performance differences to model behavior, we evaluate every model in the same deliberately minimal single-agent environment. The scaffold provides six generic tools covering shell execution, bounded file discovery and inspection, string-based editing, and termination, without any task-specific localization or repair module. Each agent receives the same fixed $V_1$ skill set, official release patch, and standard operating procedure (SOP) prompt, and edits the workspace to produce $\mathcal{S}_{V_2}$. All runs use a pre-specified turn budget. The budget is stated at initialization, and the remaining budget is reported to the agent throughout execution. If the budget is exhausted before the agent finishes, the resulting partial skill set is still evaluated. This controlled scaffold enables comparison of localization, editing, and convergence behavior under the same external constraints.

\subsection{Evaluation Metric}
\label{sec:metric}

The headline evaluation is a strictly patch-grounded removal metric. It scores whether the agent removes or modifies the $V_1$ knowledge that the release patch directly shows to be obsolete, while penalizing edits to $V_1$ lines outside the patch-supported obsolete set. Since valid $V_2$ content may admit multiple realizations, additions and reformulations in the updated skill set are assessed by the complementary NL Judge described below.

\paragraph{Gold obsolete set.} For each transition, we construct a \textbf{gold obsolete set} $G_{\Delta}$ containing the $V_1$ skill lines that the official $V_1\!\rightarrow\!V_2$ release patch directly shows to be no longer valid under $V_2$. This includes content invalidated by changes to APIs, signatures, defaults, files, commands, configurations, schemas, behavioral contracts, or other repository behavior exposed by the patch. Obsolete content may appear in prose, examples, reference tables, or scripts. Throughout the metric, a line denotes one physical line of a skill file, and diff matches are made against the original $V_1$ lines without merging, splitting, or normalization. $G_{\Delta}$ is defined on the $V_1$ side of the transition: it identifies lines that require deletion or revision while leaving the form of a valid $V_2$ replacement unconstrained. We construct $G_{\Delta}$ by automated candidate discovery followed by line-level manual verification against direct patch evidence (Appendix~\ref{app:gold-labels}).

\paragraph{Removal precision, recall, and $F_1$.} We evaluate the $V_1$-side removal target defined by $G_{\Delta}$. Let $R$ denote the set of $V_1$ skill lines that appear on the deletion side of the skill-set diff between $\mathcal{S}_{V_1}$ and the agent-produced $\mathcal{S}_{V_2}$. Thus, $R$ captures both deletions and revisions to existing guidance, while newly added $V_2$ lines are not included. A gold obsolete line counts as hit when its original form appears on the deletion side of the $\mathcal{S}_{V_1}\!\rightarrow\!\mathcal{S}_{V_2}$ diff. We deliberately do not match newly generated ``+'' lines, since a valid post-release replacement may admit multiple semantically equivalent realizations, and final-state quality is instead assessed by the NL Judge. We compute

\[
\mathrm{Recall} =
\frac{|G_{\Delta}\cap R|}{|G_{\Delta}|},
\qquad
\mathrm{Precision} =
\frac{|G_{\Delta}\cap R|}{|R|},
\qquad
F_1 =
\frac{2|G_{\Delta}\cap R|}{|G_{\Delta}|+|R|}.
\]

Recall measures the fraction of patch-verified stale lines that are removed or revised, whereas precision measures the fraction of changed $V_1$ lines that belong to $G_{\Delta}$. Their harmonic mean balances stale-content coverage against edit restraint under the minimal-edit objective. All 105 transitions have $|G_{\Delta}|>0$, so recall is always defined. When $R\neq\varnothing$ but $G_{\Delta}\cap R=\varnothing$, precision is zero, and when $R=\varnothing$, recall and $F_1$ are zero while precision is undefined.

For each transition and metric, avg@3 is the mean over three runs and best@3 is the maximum over the three runs, with undefined precision values omitted. We then macro-average these transition-level scores across the corpus. Recall and $F_1$ therefore include all 105 transitions, whereas macro precision excludes a transition only when precision is undefined in all three runs. Our headline score is avg@3 macro $F_1$, with macro recall and precision reported separately. \paragraph{Complementary NL Judge.} Because the patch-grounded metric scores the $V_1$-side removal target, the semantic correctness of replacement and newly added $V_2$ content falls to a complementary NL Judge~\citep{geval,llm_judge,prometheus}. For each run, the judge reads the release patch, the $V_1$ skills, the produced $\mathcal{S}_{V_2}$ skill set, and limited run metadata. The judge assigns five 0--2 scores. \textsc{API} checks whether the produced skills describe the correct $V_2$ API, \textsc{Files} checks whether the skill prose, scripts, and references are mutually consistent, \textsc{Info} checks whether still-valid $V_1$ guidance was retained, \textsc{Prompt} checks adherence to the patch-grounded maintenance instructions, and \textsc{NL} checks whether the prose describes the post-release state directly. API correctness, file consistency, and retention are aggregated across impacted skills, while prompt adherence and NL correctness are package-level judgments. The Sum score (0--10) is the total over these five dimensions, with structurally invalid skill sets form-gated to zero.

\section{Experiments}
\label{sec:experiments}

Our experiments center on whether agents can maintain repository skills across releases. We first conduct a focused utility study to assess whether the fixed $V_1$ skills encode repository-specific knowledge with meaningful downstream utility, thereby motivating their maintenance. We then evaluate maintenance performance and diagnose the sources of failure.

\subsection{Setup}
\label{sec:setup}

\textbf{Maintenance data.} Repo2Skill-Evo contains 105 selected official release transitions from 57 public GitHub repositories, with one fixed $V_1$ skill set per transition and 1{,}158 individual skills in total. Each transition pairs official $V_1$ and $V_2$ tags with the corresponding source diff. The corpus focuses on the fast-moving AI/ML and LLM-agent ecosystem, covering agent and RAG frameworks, training and inference stacks, and supporting infrastructure, while retaining several mature general-purpose libraries for breadth. We prioritize transitions with visible source-level changes that may affect repository-specific procedural knowledge (Appendix~\ref{app:dataset-construction}). Across the corpus, the 105 gold obsolete sets contain 12{,}217 patch-verified stale $V_1$ skill lines, with individual sets ranging from 5 to 375 lines and a median of 92 (Appendix~\ref{app:gold-size}).

\textbf{Models and protocol.} We evaluate Claude-opus-4.6, GLM-5.1~\citep{glm5}, GPT-5.4, Kimi-K2.5~\citep{kimi_k25}, Doubao-Seed2-pro~\citep{doubao_seed2}, and MiniMax-M2.5~\citep{minimax_m2}. Each model runs three times per transition under the same agent scaffold and 50-turn budget. For each transition, all runs receive the same fixed $V_1$ skill set and official $V_1\!\rightarrow\!V_2$ release patch in a standardized tool environment.

\begin{figure}[!ht]
\centering
\includegraphics[width=0.8\textwidth]{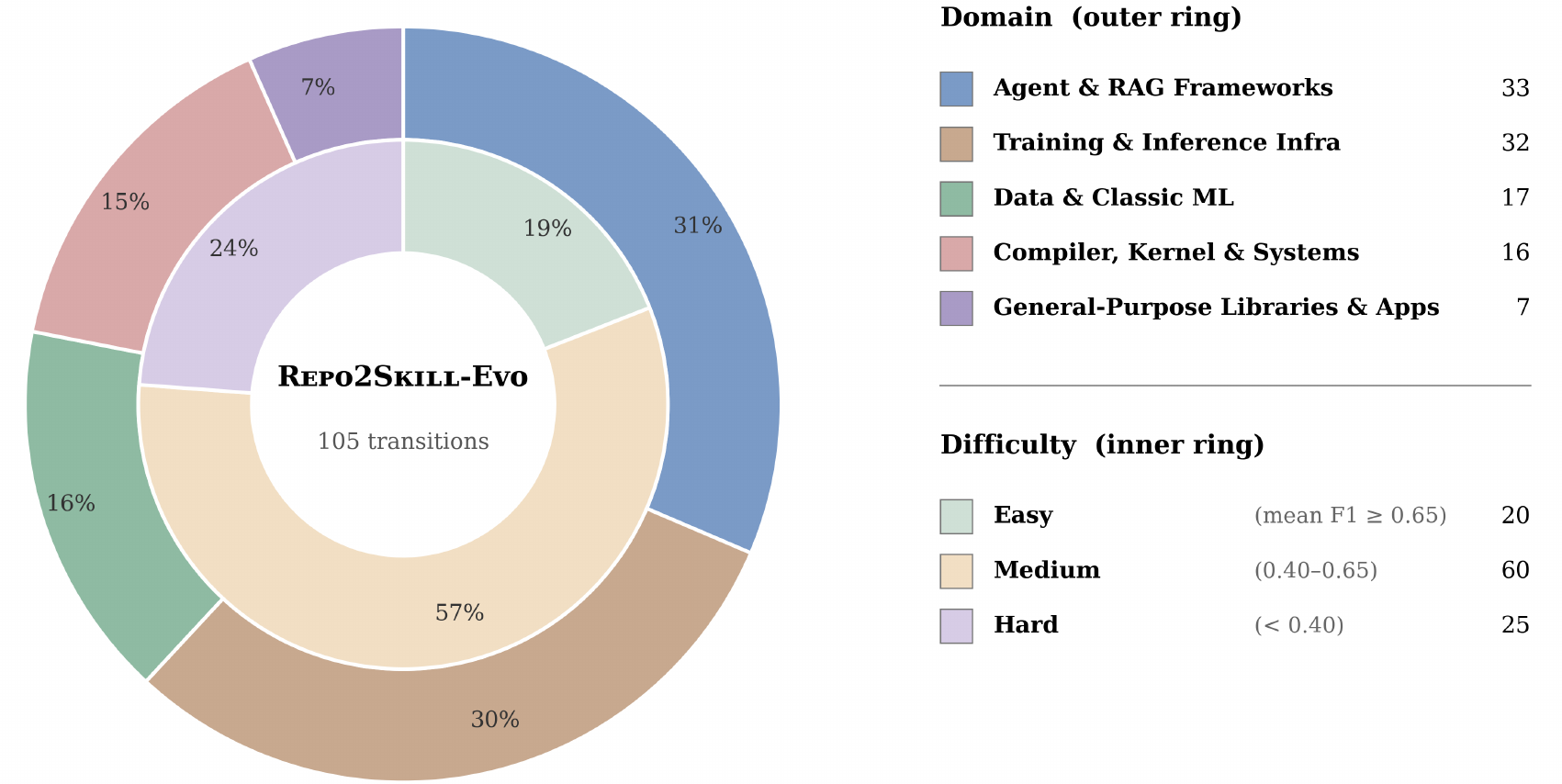}
\caption{Composition of the Repo2Skill-Evo corpus, covering 105 release transitions across 57 repositories. The outer ring shows repository domain, and the inner ring shows observed maintenance difficulty, defined by the mean avg@3 $F_1$ across the six maintenance agents. Per-transition assignments and the full score matrix appear in Appendix~\ref{app:corpus}.}
\label{fig:corpus}

\end{figure}

\textbf{Corpus composition.} Figure~\ref{fig:corpus} summarizes the corpus along two axes: repository domain and observed maintenance difficulty. By domain, the corpus includes 33 transitions from agent and RAG frameworks, 32 from training and inference infrastructure, 17 from data and classic ML libraries, 16 from compiler, kernel, and systems software, and 7 from general-purpose libraries and applications. For each transition, we average avg@3 $F_1$ across the six maintenance models and classify the resulting observed difficulty as \textsc{Easy} ($\geq 0.65$), \textsc{Medium} ($[0.40, 0.65)$), or \textsc{Hard} ($<0.40$). Appendix~\ref{app:corpus} provides the classification criteria, per-transition assignments, per-model scores, and domain-level outcomes.

\subsection{Motivating Study: Fixed Skills Provide Meaningful Repository Utility}
\label{sec:rq1}

Before evaluating whether repository skills can be maintained across releases, we first ask whether the fixed $V_1$ skills encode repository-specific knowledge worth maintaining. Prior work has shown that skill benefits vary substantially across tasks~\citep{skillsbench,swe_skills_bench}. We therefore evaluate the downstream utility of the Repo2Skill-Evo skills and examine where their gains are largest.

\paragraph{Experimental protocol.} We randomly sample ten repositories from the corpus and define one artifact-generation task per repository, together with a repository-specific utility rubric that remains hidden from the model. We evaluate GPT-5.4 in a $2\times2$ ablation over source and skill access, with three runs in each of the four conditions \texttt{baseline}, \texttt{skill-only}, \texttt{source-only}, and \texttt{source+skill}. Appendix~\ref{app:skill-gain} defines these conditions along with the three contrasts and reports the per-repository scores. We use baseline utility as a coarse proxy for prior repository knowledge and split the repositories at the median into low- and high-baseline groups (Table~\ref{tab:skill-gain}).

\begin{table}[H]
\centering
\caption{Repository-skill utility by baseline-utility group. Values are group means of artifact utility scores (0--10), manually scored under repository-specific rubrics, with three runs per repository and condition. Groups are formed by a median split of baseline utility. Skill and source deltas are relative to baseline; skill~$\Delta_{\mathrm{src}}$ is relative to source-only. Appendix~\ref{app:skill-gain} provides per-repository scores and cost statistics.}
\label{tab:skill-gain}

\scriptsize
\setlength{\tabcolsep}{3.6pt}
\begin{tabular}{lrrrrrrrr}
\toprule
Group & \#repos & baseline & skill-only & source-only & source+skill &
skill~$\Delta$ & source~$\Delta$ & skill~$\Delta_{\mathrm{src}}$ \\
\midrule
All repositories & 10 & 5.88 & 8.68 & 8.64 & 9.01 & +2.80 & +2.76 & +0.37 \\
Low-baseline repositories & 5 & 4.48 & 8.46 & 8.46 & 8.86 &
\textbf{+3.98} & +3.98 & +0.40 \\
High-baseline repositories & 5 & 7.28 & 8.89 & 8.82 & 9.15 &
+1.61 & +1.54 & +0.33 \\
\bottomrule
\end{tabular}
\end{table}

\textbf{Skills provide substantial downstream utility.} Across all ten repositories, skill-only raises mean utility from 5.88 to 8.68, a gain of 2.80 points, and nearly matches source-only at 8.64. The fixed skills therefore provide task-relevant repository knowledge that the model does not consistently recover without external context.

\textbf{Gains are largest at low baseline utility.} Skill-only raises mean utility in the low-baseline group from 4.48 to 8.46, a gain of 3.98 points, compared with 1.61 points in the high-baseline group. Within this sample, skills provide the largest absolute gains on repositories where the model performs worst without source or skill access.

\textbf{Skills provide comparable utility at substantially lower cost.} Skill-only and source-only achieve nearly identical mean utility (8.68 versus 8.64), but skill-only uses 51{,}821 mean tokens and 3.9 iterations, compared with 272{,}620 tokens and 10.8 iterations for source-only. When source is already available, adding skills increases mean utility from 8.64 to 9.01 while modestly reducing mean token usage (Appendix~\ref{app:skill-gain}). Together, these results show that the fixed $V_1$ skills encode useful, compact repository knowledge, motivating their maintenance across releases.

\subsection{Agents Do Not Reliably Maintain Repository Skills across Releases}

We now evaluate the maintenance task defined in Section~\ref{sec:evo-task}. Table~\ref{tab:main} reports the headline patch-grounded removal scores. The metric rewards the removal or revision of obsolete $V_1$ lines while penalizing edits to $V_1$ lines outside $G_{\Delta}$.

\textbf{Maintenance remains far from solved.} Claude-opus-4.6 achieves the highest avg@3 macro $F_1$ at 69.7\%, followed by GLM-5.1 at 64.3\%, while the remaining four models fall below 60\%. GPT-5.4 obtains the highest avg@3 macro recall at 74.6\%, but its substantially lower macro precision of 55.7\% limits its avg@3 macro $F_1$ to 58.8\%. Kimi-K2.5, Doubao-Seed2-pro, and MiniMax-M2.5 achieve avg@3 macro $F_1$ scores of only 47.0\%, 39.8\%, and 29.9\%, respectively.

These score profiles have direct maintenance consequences. Low recall leaves patch-verified stale guidance in the produced skill set, whereas low precision reflects broader editing outside the patch-grounded obsolete set and therefore weaker adherence to the task's minimal-edit objective. Repository-level macro averaging preserves the model ordering, and 95\% paired repository-clustered bootstrap intervals for the three adjacent avg@3 $F_1$ differences among the top four models all exclude zero (Appendix~\ref{app:robustness}).

\begin{table}[H]
\centering
\caption{Patch-grounded maintenance scores (\%) over 105 release transitions and three runs per model. R, P, and $F_1$ denote transition-macro recall, precision, and $F_1$. avg@3 averages the three runs for each transition, whereas best@3 takes the per-transition maximum over the three runs before macro-averaging. Macro $F_1$ is averaged directly over transitions and is therefore not the harmonic mean of the displayed macro R and P. Models are sorted by avg@3 macro $F_1$; the best result in each column is shown in bold.}
\label{tab:main}

\footnotesize
\setlength{\tabcolsep}{3.2pt}
\begin{tabular}{l rrr rrr}
\toprule
& \multicolumn{3}{c}{avg@3 macro} & \multicolumn{3}{c}{best@3 macro} \\
\cmidrule(lr){2-4}\cmidrule(lr){5-7}
Model & R & P & $F_1$ & R & P & $F_1$ \\
\midrule
Claude-opus-4.6  & 70.4 & \textbf{75.7} & \textbf{69.7} & 79.3 & 82.2 & \textbf{76.1} \\
GLM-5.1          & 64.1 & 72.4 & 64.3 & 78.1 & 80.9 & 73.9 \\
GPT-5.4          & \textbf{74.6} & 55.7 & 58.8 & \textbf{87.9} & 67.9 & 70.4 \\
Kimi-K2.5        & 40.4 & 71.6 & 47.0 & 55.0 & 84.9 & 60.7 \\
Doubao-Seed2-pro & 31.9 & 73.3 & 39.8 & 44.7 & 85.0 & 53.4 \\
MiniMax-M2.5     & 22.6 & 74.9 & 29.9 & 36.6 & \textbf{86.8} & 46.0 \\
\bottomrule
\end{tabular}
\end{table}

\textbf{Models occupy different points on the recall--precision tradeoff.} GPT-5.4 removes or revises a broader set of $V_1$ lines and achieves the highest recall, but many of these changes fall outside $G_{\Delta}$. Claude-opus-4.6 and GLM-5.1 maintain more balanced recall--precision profiles. The lower-scoring models tend toward high precision and low recall: MiniMax-M2.5, for example, reaches 74.9\% precision but removes or revises only 22.6\% of the verified stale content. High precision in this regime does not imply successful maintenance; it coexists with substantial under-editing. Section~\ref{sec:failure-diagnosis} connects these outcome profiles to affected-file coverage and edit-selection diagnostics derived from the execution traces and output diffs.

\textbf{Repeated attempts reveal capability but not reliability.} Selecting the highest $F_1$ among three runs for each transition raises macro $F_1$ by 6.4 points for Claude-opus-4.6 and by 9.6--16.1 points for the remaining models. These gains indicate that models sometimes produce substantially better updates than their mean single-run performance suggests. However, even best@3 reaches only 76.1\% for the strongest model, and no other model exceeds 73.9\%. Moreover, best@3 selects the strongest of three observed outcomes and therefore reflects attainable performance across repeated attempts rather than single-attempt reliability. Together with the within-transition run-to-run variability reported in Appendix~\ref{app:robustness}, the generally larger best@3--avg@3 gaps for the lower-scoring models indicate that maintenance quality is unstable across repeated attempts.

\textbf{Complementary evaluation identifies the same upper tier.} Table~\ref{tab:nljudge} reports five-dimensional NL Judge scores assigned by GPT-5.4 and Claude-opus-4.6. Both judges rank Claude-opus-4.6 first, GPT-5.4 second, and GLM-5.1 third. They therefore identify the same top-three set as the patch-grounded evaluation. All models produce at least some structurally invalid skill sets, with 7--26 invalid runs out of 315 per model. The NL Judge complements the headline metric by assessing the final skill set across five rubric dimensions, whereas the patch-grounded removal metric provides a direct and auditable measure of stale-content removal and off-target editing. Full dimension-level results appear in Appendix~\ref{app:nljudge-full}.

\begin{table}[H]
\centering
\caption{Complementary NL Judge scores over 315 runs per maintenance model. Each judge reports the mean 0--10 Sum score, with structurally invalid outputs form-gated to zero before averaging. Invalid reports the number of such outputs. The best score in each judge column is shown in bold.}
\label{tab:nljudge}

\footnotesize
\setlength{\tabcolsep}{6pt}
\begin{tabular}{lrrr}
\toprule
Model & GPT-5.4 judge & Claude-opus-4.6 judge & Invalid \\
\midrule
Claude-opus-4.6  & \textbf{7.54} & \textbf{7.86} & 9 \\
GLM-5.1          & 7.09 & 6.98 & 7 \\
GPT-5.4          & 7.18 & 7.31 & 26 \\
Kimi-K2.5        & 6.49 & 6.24 & 14 \\
Doubao-Seed2-pro & 6.54 & 6.22 & 15 \\
MiniMax-M2.5     & 6.01 & 5.27 & 8 \\
\bottomrule
\end{tabular}
\end{table}

\textbf{The maintenance shortfall extends across the corpus.} Transition difficulty varies substantially, but poor performance is not confined to a few isolated cases. Under the observed-difficulty stratification defined in Section~\ref{sec:setup}, 25 of the 105 transitions fall below 0.40, 60 fall in $[0.40,0.65)$, and only 20 reach at least 0.65 when averaging the six models' avg@3 $F_1$ scores. Thus, 85 transitions fall below the \textsc{Easy} threshold. Appendix~\ref{app:corpus} provides the full stratification and representative high- and low-scoring transitions.

\subsection{Failure Diagnosis: Coverage and Editing Bottlenecks}
\label{sec:failure-diagnosis}

Maintenance can fail because agents miss affected skill files, edit too little or too broadly, or run out of turns before finishing. We analyze all 1{,}890 runs using execution traces and maintained-output diffs. Table~\ref{tab:behavior} summarizes the resulting model-level diagnostics. Affected-file coverage measures the fraction of gold-affected skill files localized during a run. Removal load, $|R|/|G_{\Delta}|$, measures the volume of changed $V_1$ lines relative to the gold removal target, while off-target editing measures the fraction of changed $V_1$ lines outside $G_{\Delta}$. Together, these diagnostics distinguish missed localization and under-editing from broad, poorly targeted editing. They are used only for post hoc analysis. Appendix~\ref{app:behavior-metrics} provides the full definitions.

\begin{table}[H]
\centering
\caption{Post hoc diagnostics over 315 runs per model. Coverage and removal load are mean ratios; full coverage, off-target editing, no-removal runs, and budget exhaustion are percentages; turns is the mean number of turns per run.}
\label{tab:behavior}

\scriptsize
\setlength{\tabcolsep}{3.5pt}
\begin{tabular}{l rrrrrrr}
\toprule
Model & mean file cov. & full file cov. & removal load & off-target &
no-removal & budget exh. & turns \\
\midrule
Claude-opus-4.6  & 0.95 & 75.9 & 1.01 & 24.0 & 0.6 & 1.9 & 25.3 \\
GLM-5.1          & 0.95 & 78.1 & 1.01 & 27.6 & 0.6 & 41.3 & 42.4 \\
GPT-5.4          & 0.93 & 69.2 & 2.84 & 44.3 & 0.0 & 0.0 & 17.8 \\
Kimi-K2.5        & 0.74 & 25.7 & 0.60 & 28.6 & 2.9 & 47.3 & 43.5 \\
Doubao-Seed2-pro & 0.53 & 10.5 & 0.42 & 25.9 & 4.1 & 14.6 & 33.1 \\
MiniMax-M2.5     & 0.51 & 9.5  & 0.30 & 22.7 & 20.0 & 68.6 & 47.0 \\
\bottomrule
\end{tabular}
\end{table}

\textbf{Coverage is strongly associated with maintenance quality.} Across all 1{,}890 runs, affected-file coverage correlates with both $F_1$ ($r=0.650$) and recall ($r=0.701$), and these associations persist after removing additive model and transition effects (Appendix~\ref{app:behavior-metrics}). Runs with full affected-file coverage average 67.3\% $F_1$ and 72.1\% recall, compared with 41.0\% $F_1$ for partial-coverage runs. These results consistently associate incomplete localization with missed stale content.

\textbf{Oracle localization improves performance but leaves substantial residual errors.} We test localization more directly through an oracle ablation on the hardest-20 subset, restricted to transitions with at least 20 gold obsolete lines. We append only the gold-affected skill roots and file paths to the original prompt, without providing patch hunks, line numbers, obsolete-line labels, or edit instructions.

\begin{table}[H]
\centering
\caption{Oracle skill-file localization on the hardest-20 subset. Values are avg@3 macro scores (\%). Gain columns report mean per-transition oracle-minus-baseline differences.}
\label{tab:oracle-localization}

\scriptsize
\setlength{\tabcolsep}{2.4pt}
\begin{tabular}{@{}lrrr rrr rrr@{}}
\toprule
& \multicolumn{3}{c}{Baseline} & \multicolumn{3}{c}{Oracle} &
\multicolumn{3}{c}{Gain} \\
\cmidrule(lr){2-4}\cmidrule(lr){5-7}\cmidrule(l){8-10}
Model & R & P & $F_1$ & R & P & $F_1$ &
$\Delta$R & $\Delta$P & $\Delta F_1$ \\
\midrule
Claude-opus-4.6  & 55.4 & 64.4 & 53.6 & 63.5 & 68.2 & 62.3 & +8.1  & +3.8  & +8.7 \\
GLM-5.1          & 46.2 & 65.2 & 48.7 & 54.8 & 66.0 & 55.9 & +8.6  & +0.7  & +7.2 \\
GPT-5.4          & 65.6 & 47.9 & 48.2 & 73.3 & 50.5 & 54.6 & +7.7  & +2.6  & +6.3 \\
Kimi-K2.5        & 14.2 & 50.4 & 19.0 & 31.8 & 61.9 & 36.6 & +17.6 & +11.5 & +17.6 \\
Doubao-Seed2-pro & 11.5 & 58.6 & 16.9 & 30.2 & 68.9 & 37.6 & +18.7 & +10.3 & +20.8 \\
MiniMax-M2.5     & 6.7  & 50.0 & 10.1 & 18.0 & 69.4 & 23.0 & +11.2 & +19.1 & +12.9 \\
\midrule
\textbf{Mean}    & 33.3 & 56.1 & 32.8 & 45.3 & 64.1 & 45.0 & +12.0 & +8.0 & +12.2 \\
\bottomrule
\end{tabular}
\end{table}

Averaged across the six models, avg@3 macro $F_1$ increases from 32.8\% to 45.0\%, a mean transition-level gain of 12.2 points (95\% paired bootstrap CI: $[+6.8,+17.8]$), while recall increases by 12.0 points. Recall improves for every model. Yet even with oracle paths, the best model reaches only 62.3\% $F_1$ on this subset. File-level localization is therefore important but insufficient: agents must still identify the affected claims within each file and select the appropriate edits. Appendix~\ref{app:oracle-localization} provides the full experimental details.

\begin{figure}[!ht]
\centering
\includegraphics[width=0.6\linewidth]{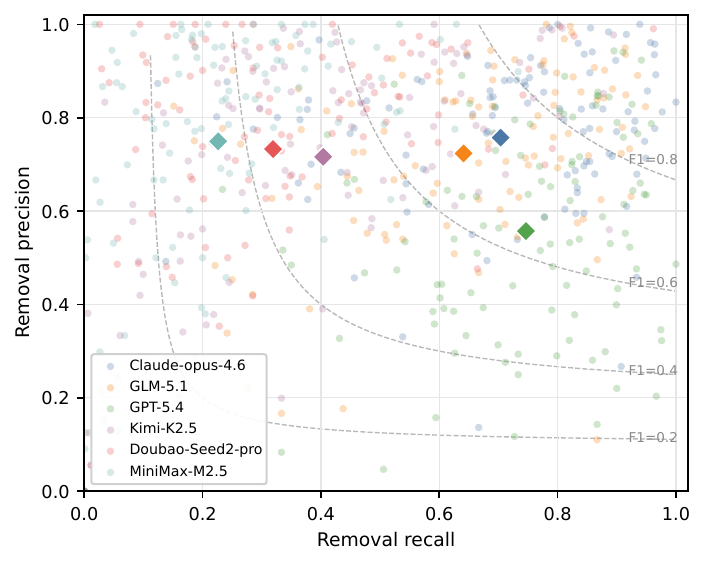}
\caption{Removal recall versus precision for each model--transition pair after avg@3 aggregation. Pairs with undefined precision are omitted. Diamonds mark per-model means, and light curves show iso-$F_1$ contours.}
\label{fig:scatter}

\end{figure}

\textbf{Models exhibit distinct failure profiles.} Figure~\ref{fig:scatter} visualizes the transition-level recall--precision tradeoff behind the model averages.

The diagnostics reveal different routes to failure. GPT-5.4 achieves high affected-file coverage but has a removal load of 2.84 and the highest off-target editing rate, consistent with its high-recall, low-precision profile. In contrast, Kimi-K2.5, Doubao-Seed2-pro, and MiniMax-M2.5 show substantially lower coverage and removal loads, indicating that many failures arise before enough affected content is reached and edited. Kimi-K2.5 and MiniMax-M2.5 also frequently exhaust the turn budget. GLM-5.1 achieves high coverage and a near-unit removal load but exhausts the budget in 41.3\% of runs, showing that finding the affected files does not by itself ensure successful completion. Claude-opus-4.6 exhibits the most balanced profile, combining high coverage, a near-unit removal load, and infrequent budget exhaustion. Overall, the results point to two complementary maintenance bottlenecks: localizing the affected skill content and selecting edits that are sufficiently complete without becoming overly broad.

\section{Limitations}
\label{sec:limitations}

Our dataset is a curated staleness-positive challenge set, so our results characterize maintenance difficulty on releases known to affect skills, with the prevalence of staleness across releases forming a separate question (Appendix~\ref{app:dataset-construction}). Our evaluation measures patch-grounded maintenance fidelity, leaving the downstream utility of maintained skills to future work. Although the NL Judge provides a complementary assessment of final-state quality, we do not rerun the maintained skill sets on downstream repository tasks.

\section{Conclusion}

Repository-grounded skills turn repository-specific procedures into reusable external knowledge, but their validity does not persist automatically. As repositories evolve, previously correct guidance can become misleading while remaining loadable and retrievable. We call this release-driven failure mode \emph{silent staleness}.

Repo2Skill-Evo operationalizes this lifecycle problem as a measurable maintenance task by pairing fixed $V_1$ skill sets with official release patches and evaluating the resulting updates against patch-grounded targets. Across 57 repositories and 105 selected release transitions, even the strongest of the six evaluated agents reaches only 69.7\% avg@3 macro $F_1$. Trace and output analyses characterize two complementary maintenance bottlenecks: incomplete localization of affected skill files and imperfect edit selection. Oracle skill-file localization improves performance but leaves substantial residual errors, indicating that localization alone is insufficient. Together, these findings suggest that repository skills should be treated as versioned knowledge assets that carry the release they were distilled from and require maintenance as the repository evolves.

\clearpage
\section{Contributions}
\textbf{Project Lead}(${\alpha-\beta}$ order)

Chenyuan Duan (duanchenyuan26@stu.pku.edu.cn)

Ge Shi(23121732@bjtu.edu.cn)\\

\textbf{Core Contributors}(${\alpha-\beta}$ order)

Zineng Mao, Ge Zhang\\

\textbf{Contributors}(${\alpha-\beta}$ order)

Hao Liang, Yinzhu Piao, Yuchen Wu, Zhixin Yao\\

\textbf{Corresponding}(${\alpha-\beta}$ order)

Kaiyu Huang, Wenhao Huang, Linzhuang Sun

Shen Yan(sheny@bytedance.com)

Wentao Zhang(wentao.zhang@pku.edu.cn)

\clearpage
\bibliographystyle{plainnat}
\bibliography{main}

\clearpage

\beginappendix
\section{Skill-Gain Utility Study}
\label{app:skill-gain}

Section~\ref{sec:rq1} reports the aggregate findings of the motivating utility study. Here we provide the experimental protocol, per-repository results, baseline-utility grouping procedure, and execution costs.

\paragraph{Experimental protocol.} Each of the ten sampled repositories is represented once. GPT-5.4 performs three runs in each cell of a $2\times2$ ablation over access to the fixed $V_1$ skill set and the $V_1$ repository checkout:

\begin{itemize}

\item \textbf{baseline}: neither source nor skill is available;

\item \textbf{skill-only}: the skill set is available and the source is unavailable;

\item \textbf{source-only}: the source is available and the skill set is unavailable;

\item \textbf{source+skill}: both the source and the skill set are available.

\end{itemize}

The four conditions use the same repository-specific task, evaluation rubric, agent scaffold, and execution budget, differing only in source and skill access. Skill~$\Delta$ is \texttt{skill-only} minus \texttt{baseline}, source~$\Delta$ is \texttt{source-only} minus \texttt{baseline}, and skill~$\Delta_{\mathrm{src}}$ is \texttt{source+skill} minus \texttt{source-only}. These contrasts measure, respectively, the utility of skills without source access, the utility of source access without skills, and the incremental utility of skills when source is already available. Table~\ref{tab:skill-gain} reports grouped means over the same per-repository scores shown in Table~\ref{tab:rq1-per-repo}.

\begin{table}[htbp]
\centering
\caption{Per-repository artifact utility scores for the ten-repository motivating study. Scores are 0--10 means over three runs. Skill~$\Delta$ is \texttt{skill-only} minus \texttt{baseline}, source~$\Delta$ is \texttt{source-only} minus \texttt{baseline}, and skill~$\Delta_{\mathrm{src}}$ is \texttt{source+skill} minus \texttt{source-only}. Rows are ordered by baseline utility, with the median split separating the low- and high-baseline groups.}
\label{tab:rq1-per-repo}

\scriptsize
\setlength{\tabcolsep}{2.7pt}
\begin{tabular}{lrrrrrrr}
\toprule
Repository & baseline & skill-only & source-only & source+skill &
skill~$\Delta$ & source~$\Delta$ & skill~$\Delta_{\mathrm{src}}$ \\
\midrule
\texttt{agentmemory} & 3.43 & 8.57 & 8.83 & 9.07 & +5.13 & +5.40 & +0.23 \\
\texttt{oh-my-pi} & 4.33 & 8.73 & 7.90 & 9.10 & +4.40 & +3.57 & +1.20 \\
\texttt{harbor} & 4.70 & 8.20 & 8.50 & 8.63 & +3.50 & +3.80 & +0.13 \\
\texttt{deer-flow} & 4.73 & 8.47 & 9.07 & 9.17 & +3.73 & +4.33 & +0.10 \\
\texttt{hermes-agent} & 5.20 & 8.33 & 8.00 & 8.33 & +3.13 & +2.80 & +0.33 \\
\midrule
\texttt{browser-use} & 6.37 & 8.70 & 8.73 & 9.00 & +2.33 & +2.37 & +0.27 \\
\texttt{lightrag} & 6.83 & 8.67 & 9.13 & 9.17 & +1.83 & +2.30 & +0.03 \\
\texttt{mem0} & 7.17 & 8.60 & 8.67 & 9.00 & +1.43 & +1.50 & +0.33 \\
\texttt{langgraph} & 7.83 & 9.17 & 8.70 & 9.20 & +1.33 & +0.87 & +0.50 \\
\texttt{pydantic} & 8.20 & 9.33 & 8.87 & 9.40 & +1.13 & +0.67 & +0.53 \\
\midrule
\textbf{Mean} & 5.88 & 8.68 & 8.64 & 9.01 & +2.80 & +2.76 & +0.37 \\
\bottomrule
\end{tabular}
\end{table}

\paragraph{Baseline-utility grouping.} We use baseline utility as a coarse operational proxy for repository-specific prior knowledge and split the repositories at the median. The five low-baseline repositories are \texttt{agentmemory}, \texttt{oh-my-pi}, \texttt{harbor}, \texttt{deer-flow}, and \texttt{hermes-agent}. The remaining five form the high-baseline group. Group means appear in Table~\ref{tab:skill-gain}, and the per-repository values from which they are computed appear in Table~\ref{tab:rq1-per-repo}. Because baseline utility may also reflect task difficulty, and because lower-baseline tasks have more headroom on the bounded 0--10 scale, we interpret this grouping descriptively.

\paragraph{Execution cost.} Table~\ref{tab:rq1-cost} reports utility and execution cost for all four conditions. Each row aggregates 30 runs: three runs on each of ten repositories.

\begin{table}[htbp]
\centering
\caption{Artifact utility and execution cost across the four access conditions. Values are means over ten repositories and three runs per repository and condition.}
\label{tab:rq1-cost}

\footnotesize
\setlength{\tabcolsep}{8pt}
\begin{tabular}{lrrr}
\toprule
Condition & Utility & Mean tokens & Mean iterations \\
\midrule
Baseline & 5.88 & 21{,}057 & 4.4 \\
Skill-only & 8.68 & 51{,}821 & 3.9 \\
Source-only & 8.64 & 272{,}620 & 10.8 \\
Source+skill & 9.01 & 237{,}020 & 9.3 \\
\bottomrule
\end{tabular}
\end{table}

Table~\ref{tab:rq1-cost} quantifies the cost reduction noted in Section~\ref{sec:rq1}. Adding skills on top of source access reduces mean tokens from 272{,}620 to 237{,}020 and mean iterations from 10.8 to 9.3, so the +0.37-point utility gain of \texttt{source+skill} over \texttt{source-only} comes at slightly lower cost.

\section{Maintenance Dataset Construction}

Repo2Skill-Evo is constructed before any maintenance model is evaluated. The construction proceeds in four steps:
\begin{enumerate}

\item \textbf{Select release transitions.} We identify official releases with publicly available source diffs (Appendix~\ref{app:dataset-construction}).

\item \textbf{Materialize release instances.} Each transition records its repository, $V_1$ and $V_2$ version tags, commit hashes, and official $V_1\!\rightarrow\!V_2$ release patch.

\item \textbf{Construct the fixed $V_1$ skill set.} Repo2Skill produces candidate skill sets from the pre-release repository, after which an expert selects, revises, and verifies the final $\mathcal{S}_{V_1}$ used by all maintenance agents (Appendix~\ref{app:v1-skills}).

\item \textbf{Construct the gold obsolete set.} We identify the $V_1$ skill lines that the release patch directly shows to be invalid under $V_2$, producing the patch-grounded gold set $G_{\Delta}$ (Appendix~\ref{app:gold-labels}). Transitions with $G_{\Delta}=\varnothing$ are excluded, yielding the final staleness-positive corpus.

\end{enumerate}

Gold annotations are stored separately from the maintenance instances and are used only for evaluation. Maintenance agents receive neither $G_{\Delta}$ nor its annotation-side representation.

\subsection{Transition Selection and Construction}
\label{app:dataset-construction}

\paragraph{Source and scope.} Beyond the corpus-level counts given in Section~\ref{sec:setup}, each transition records the repository, the $V_1$ and $V_2$ version tags, the base and target commit hashes, and the corresponding public source diff, making every instance traceable to an upstream release. Selection favors actively maintained projects; among their releases we prefer diffs that touch interfaces, defaults, file layouts, commands, configurations, or documented repository behavior, since these are the change classes that can invalidate procedural guidance. The complete repository and release list is included in the released materials.

\paragraph{Recency and contamination.} The selected transitions span a broad range of version histories, including releases that postdate the knowledge cutoffs of some evaluated models. We do not assume that the repositories themselves were absent from model pretraining. However, the maintenance task turns on aligning a fixed $V_1$ skill set with the changes in one specific $V_1\!\rightarrow\!V_2$ release patch, so performance depends on the evidence that the patch itself supplies. We release version tags and commit hashes to support independent dating and auditing.

\subsection{Fixed $V_1$ Skill Set}
\label{app:v1-skills}

Section~\ref{sec:repo2skill} states the candidate-generation and expert-selection procedure. Here we record the two properties that matter for the validity of the benchmark. First, \emph{isolation}: both candidates are generated from the pre-release repository alone, and the expert revision draws on the same $V_1$ evidence, so the release patch and the gold obsolete labels remain outside the construction of $\mathcal{S}_{V_1}$ from end to end. Second, \emph{review scope}: the expert reviews the package as a whole, including \texttt{SKILL.md}, scripts, and references. Cited paths and symbols are checked against the repository, procedural guidance is reviewed for factual consistency, and the package must satisfy the structural and grounding requirements before it is frozen.

\subsection{Gold Obsolete Set Construction}
\label{app:gold-labels}

\paragraph{Candidate discovery and manual verification.} In the automated stage, a script extracts repository-specific interfaces referenced by the $V_1$ skill set, including symbols, paths, commands, and configuration keys, and matches them against changed lines in the release patch. This stage surfaces potentially affected skill content and is used only to assist candidate discovery. For every transition, the $V_1$ skill set is then inspected together with the corresponding release patch. Candidate lines whose documented guidance remains valid under $V_2$ are removed, while stale content missed by the automatic pass is added, including semantically expressed claims and lines whose validity depends on an affected interface or behavior. Every final line in $G_{\Delta}$ is manually verified against direct patch evidence before being used as a gold removal target.

\paragraph{Representative obsolete-content categories.} Table~\ref{tab:obsolete-categories} summarizes common sources of skill invalidation. These categories sample the range of changes observed across transitions, and membership in $G_{\Delta}$ follows from direct patch evidence for each individual line.

\begin{table}[htbp]
\centering
\caption{Representative sources of $V_1$ skill obsolescence captured by $G_{\Delta}$.}
\label{tab:obsolete-categories}

\footnotesize
\setlength{\tabcolsep}{4.5pt}
\begin{tabular}{ll}
\toprule
Repository change & Example effect on $V_1$ guidance \\
\midrule
API or interface change & Referenced functions, classes, modules, or interfaces are removed or renamed \\
Signature or default change & Arguments, defaults, return contracts, or invocation patterns become invalid \\
File or entry-point change & Referenced paths, scripts, modules, or entry points move or disappear \\
Command or configuration change & Flags, commands, configuration keys, or schemas change \\
Behavioral contract change & Previously documented behavior is contradicted by the updated implementation \\
\bottomrule
\end{tabular}
\end{table}

The resulting stale knowledge is not limited to the line that explicitly names the changed repository element. When a patch-supported change also invalidates dependent instructions, examples, table entries, or script statements, those lines are included in $G_{\Delta}$ as well.

\paragraph{Example.} Consider the transition \texttt{flask-2.2.5-2.3.0}. A $V_1$ skill, \texttt{app-env-config}, describes environment mode using \texttt{FLASK\_ENV}, the \texttt{ENV} configuration key, and the \texttt{app.env} property. The release removes this environment-mode surface, making the corresponding $V_1$ guidance obsolete. $G_{\Delta}$ therefore includes the affected reference-table rows, prose instructions that use \texttt{app.config["ENV"]}, and script lines that set \texttt{os.environ["FLASK\_ENV"]} or access \texttt{app.env}. This example illustrates how a single release change can invalidate repository-specific guidance expressed across multiple files and content forms within a skill package.

\section{Evaluation Protocol and Diagnostics}

\subsection{Behavioral Diagnostics}
\label{app:behavior-metrics}

Table~\ref{tab:behavior-defs} defines the post hoc diagnostics used in Section~\ref{sec:failure-diagnosis}. Affected-file coverage is derived from agent traces and maintained outputs, while the remaining diff-based diagnostics are computed from the $\mathcal{S}_{V_1}\!\rightarrow\!\mathcal{S}_{V_2}$ diff. A gold-affected skill file is a file in $\mathcal{S}_{V_1}$ containing at least one line in $G_{\Delta}$, and a file counts as localized when the agent directly reads it, explicitly names or targets it in the interaction trace, or modifies it in the maintained output. Localization therefore requires evidence that the agent attended to the file, which places paths appearing only in tool-returned search results outside the criterion. Affected-file coverage is the fraction of gold-affected skill files meeting this criterion.
\begin{table}[htbp]
\centering
\caption{Definitions of the post hoc diagnostics reported in Table~\ref{tab:behavior}.}
\label{tab:behavior-defs}

\footnotesize
\setlength{\tabcolsep}{4.5pt}
\begin{tabular}{@{}p{0.25\linewidth}p{0.70\linewidth}@{}}
\toprule
Metric & Definition \\
\midrule
\texttt{mean file cov.} & Run-level affected-file coverage, reported as the mean across runs. \\
\texttt{full file cov.} & Run-level indicator that every gold-affected skill file is localized, reported as the percentage of runs satisfying the indicator. \\
\texttt{removal load} & Run-level ratio $|R|/|G_{\Delta}|$, reported as the mean across runs. \\
\texttt{off-target} & Run-level fraction $|R\setminus G_{\Delta}|/|R|$, defined when $R\neq\varnothing$ and reported as the mean over defined runs. \\
\texttt{no-removal} & Run-level indicator that $R=\varnothing$, reported as the percentage of runs satisfying the indicator. \\
\texttt{budget exh.} & Run-level indicator that \texttt{run\_outcome=max\_iterations}, corresponding to 50 agent turns without a standalone \texttt{finish}, reported as the percentage of runs satisfying the indicator. \\
\texttt{turns} & Number of agent turns in a run, reported as the mean across runs. \\
\bottomrule
\end{tabular}
\end{table}

Budget exhaustion covers runs that reach the 50-turn limit while still acting on the workspace, so a run whose final turn is a standalone \texttt{finish} counts as a completed run even on turn 50. Partial outputs from budget-exhausted runs enter the evaluation on the same terms as complete ones.

\paragraph{Coverage-association robustness.} The pooled correlations reported in Section~\ref{sec:failure-diagnosis} use all 1{,}890 runs. As a robustness check, we first average the three runs for each model--transition pair, yielding 630 observations. We then residualize affected-file coverage, $F_1$, and recall with respect to additive model and transition fixed effects and compute correlations between the resulting residuals.

Coverage remains positively associated with $F_1$ ($r=0.477$) and recall ($r=0.455$). The associations are attenuated relative to the pooled run-level correlations in Section~\ref{sec:failure-diagnosis}, indicating that the relationship holds within models as well as across them.

\paragraph{Removal-load stratification.} Runs with removal load in $[0.75,1.25)$ average 77.3\% $F_1$, compared with 25.0\% for runs below 0.5. Runs with removal load at or above 1.25 achieve 80.8\% recall but 44.7\% precision. These patterns further distinguish insufficient editing from broad editing that improves stale-content coverage at the expense of edit selectivity.

\subsection{Oracle Skill-File Localization Ablation}
\label{app:oracle-localization}

The oracle ablation in Section~\ref{sec:failure-diagnosis} tests the contribution of skill-file localization more directly. We first restrict the 105 transitions to those with at least 20 gold obsolete lines. Among the eligible transitions, we rank them by the mean baseline avg@3 $F_1$ across the six maintenance models and select in ascending order, allowing at most one transition per repository, until 20 transitions are retained. This produces the hardest-20 subset without allowing multiple releases from the same repository to dominate it.

For each selected transition and model, we keep the original maintenance scaffold, tools, 50-turn budget, release patch, and fixed $V_1$ skill set unchanged, so the oracle hint described in Section~\ref{sec:failure-diagnosis} is the only intervention. Agents must still inspect the identified files, trace the relevant release changes, and determine the required updates. Each of the six models is run three times per transition, yielding 360 oracle runs.

We compare the oracle runs with the original baseline runs restricted to the same 20 transitions using the identical patch-grounded removal metric. Per-model confidence intervals are bootstrapped over the 20 transition-level $\Delta F_1$ values. For the overall interval, we first average $\Delta F_1$ across the six models within each transition and bootstrap the resulting 20 transition-level means. Baseline and oracle precision are each aggregated over transitions for which precision is defined in the corresponding condition. The reported $\Delta P$ instead averages per-transition oracle-minus-baseline differences where both values are defined, and therefore need not equal the difference between the two displayed aggregate precision values.

\subsection{Maintenance and Judge Instructions}
\label{app:prompts}

We release the full instructions with the accompanying materials. Baseline maintenance agents receive the fixed $V_1$ skill set and official release patch alone, with gold-affected paths supplied only in the oracle localization ablation of Appendix~\ref{app:oracle-localization}.

\paragraph{Maintenance instruction.} The maintenance agent is instructed to read the existing skill set and official release patch, make minimal patch-grounded updates, cover affected prose, scripts, and references, and describe the post-release state directly rather than narrating the migration. It should preserve unaffected package structure unless the release requires a change, and it should finish within the fixed 50-turn budget. The total and remaining turn budget are reported throughout execution.

\paragraph{NL Judge instruction.} The NL Judge works from the official release patch, the fixed $V_1$ skill set, the produced $V_2$ skill set, and limited run metadata alone, so its assessment stays independent of $G_{\Delta}$ and the patch-grounded removal scores. The full rubric is included in the released materials.

\section{Additional Results and Corpus Statistics}

\subsection{Full NL Judge Results}
\label{app:nljudge-full}

Table~\ref{tab:nljudge-full} reports the five dimension-level scores underlying the NL Judge results in Table~\ref{tab:nljudge}. Form-gated outputs contribute zero to each dimension and to the Sum score. Values are averaged over all 315 runs for each maintenance model.
\begin{table}[htbp]
\centering
\caption{ Full NL Judge results. Sum is the 0--10 total over five 0--2 dimensions: \textsc{API} correctness, \textsc{Files} consistency, \textsc{Info} retention, \textsc{Prompt} adherence, and \textsc{NL} current-state prose correctness. Invalid reports the number of form-gated runs.}
\label{tab:nljudge-full}

\scriptsize
\setlength{\tabcolsep}{3.0pt}
\begin{tabular}{@{}lrrrrrrr@{}}
\toprule
& \multicolumn{5}{c}{Dimension scores (0--2)}
& \multicolumn{1}{c}{Total}
& \multicolumn{1}{c}{Count} \\
\cmidrule(lr){2-6}\cmidrule(lr){7-7}\cmidrule(l){8-8}
Model & API & Files & Info & Prompt & NL & Sum & Invalid \\
\midrule
\multicolumn{8}{@{}l}{\textit{GPT-5.4 judge}} \\
Claude-opus-4.6  & 1.38 & 1.00 & 1.75 & 1.65 & 1.76 & \textbf{7.54} & 9  \\
GLM-5.1          & 1.20 & 1.03 & 1.76 & 1.37 & 1.74 & 7.09 & 7  \\
GPT-5.4          & 1.25 & 1.04 & 1.52 & 1.66 & 1.70 & 7.18 & 26 \\
Kimi-K2.5        & 1.06 & 0.92 & 1.74 & 1.20 & 1.57 & 6.49 & 14 \\
Doubao-Seed2-pro & 1.05 & 0.96 & 1.81 & 1.23 & 1.49 & 6.54 & 15 \\
MiniMax-M2.5     & 0.93 & 1.02 & 1.86 & 0.92 & 1.28 & 6.01 & 8  \\
\addlinespace[3pt]
\multicolumn{8}{@{}l}{\textit{Claude-opus-4.6 judge}} \\
Claude-opus-4.6  & 1.38 & 1.24 & 1.60 & 1.84 & 1.81 & \textbf{7.86} & 9  \\
GLM-5.1          & 1.20 & 1.09 & 1.61 & 1.36 & 1.72 & 6.98 & 7  \\
GPT-5.4          & 1.24 & 1.26 & 1.32 & 1.76 & 1.73 & 7.31 & 26 \\
Kimi-K2.5        & 0.99 & 0.89 & 1.53 & 1.24 & 1.59 & 6.24 & 14 \\
Doubao-Seed2-pro & 0.94 & 0.90 & 1.54 & 1.33 & 1.52 & 6.22 & 15 \\
MiniMax-M2.5     & 0.86 & 0.79 & 1.53 & 0.86 & 1.23 & 5.27 & 8  \\
\bottomrule
\end{tabular}
\end{table}

\subsection{Statistical Robustness of the Main Results}
\label{app:robustness}

The headline metric gives equal weight to the 105 release transitions, but 28 of the 57 repositories contribute more than one transition. We therefore examine robustness to repository-level clustering, alternative repository-level aggregation, and run-to-run variability.

\paragraph{Repository-clustered confidence intervals.} We resample the 57 repositories with replacement, retaining all transitions from each sampled repository, and recompute avg@3 macro $F_1$ over 10{,}000 bootstrap samples. Table~\ref{tab:robustness} reports the resulting percentile 95\% confidence intervals.

\paragraph{Repository-level macro average.} As an alternative aggregation, we first average transition-level avg@3 $F_1$ within each repository and then average across the 57 repositories. This removes the additional weight given to repositories that contribute multiple transitions. The resulting repository-level macro scores preserve the model ordering and differ from the transition-level scores by only $+0.9$ to $+3.4$ points.

\begin{table}[htbp]
\centering
\caption{Robustness of the headline avg@3 macro $F_1$ (\%). ``Transition-macro'' is the main-text aggregation with equal weight per transition; the 95\% CI is obtained by repository-clustered bootstrap. ``Repository-macro'' averages transition-level scores within each repository before averaging across repositories.}
\label{tab:robustness}

\footnotesize
\setlength{\tabcolsep}{6pt}
\begin{tabular}{lrrr}
\toprule
Model & Transition-macro & 95\% CI (cluster) & Repository-macro \\
\midrule
Claude-opus-4.6  & 69.7 & [65.7, 73.8] & 72.7 \\
GLM-5.1          & 64.3 & [60.5, 68.2] & 67.2 \\
GPT-5.4          & 58.8 & [55.1, 62.8] & 62.3 \\
Kimi-K2.5        & 47.0 & [42.8, 51.3] & 49.0 \\
Doubao-Seed2-pro & 39.8 & [35.3, 44.8] & 40.7 \\
MiniMax-M2.5     & 29.9 & [26.0, 34.0] & 31.8 \\
\bottomrule
\end{tabular}
\end{table}

\paragraph{Paired repository-clustered bootstrap.} We further assess the three adjacent differences among the top four models using the same repository-level resampling while pairing scores within each transition. All three 95\% confidence intervals exclude zero: Claude-opus-4.6 exceeds GLM-5.1 by $5.4$ points ($[+2.8,+8.0]$), GLM-5.1 exceeds GPT-5.4 by $5.4$ points ($[+2.0,+8.7]$), and GPT-5.4 exceeds Kimi-K2.5 by $11.9$ points ($[+8.1,+15.8]$).

\paragraph{Run-to-run variance.} For each model and transition, we compute the standard deviation of $F_1$ across the three runs and then average over transitions. The resulting values are 6.1 for Claude-opus-4.6, 9.0 for GLM-5.1, 11.6 for GPT-5.4, 12.0 for Kimi-K2.5, 11.9 for Doubao-Seed2-pro, and 14.7 for MiniMax-M2.5. Run-to-run variability is generally larger for the lower-scoring models, providing a complementary view of maintenance reliability alongside avg@3 and best@3.

\subsection{Obsolete-Set Size Distribution}
\label{app:gold-size}

Table~\ref{tab:gold-size} summarizes the size of the patch-verified obsolete set $G_{\Delta}$ across the 105 release transitions. The corpus contains 12{,}217 obsolete $V_1$ skill lines in total, ranging from 5 to 375 per transition, with a mean of 116.4 and quartiles of 46, 92, and 156. Only 7 transitions contain fewer than 20 obsolete lines, whereas 48 contain at least 100. Thus, most selected transitions require updating a nontrivial amount of skill content.

\begin{table}[htbp]
\centering
\caption{Distribution of the patch-verified obsolete-set size $|G_{\Delta}|$ over the 105 release transitions.}
\label{tab:gold-size}

\footnotesize
\setlength{\tabcolsep}{6pt}
\begin{tabular}{lrr}
\toprule
$|G_{\Delta}|$ (obsolete $V_1$ skill lines) & Transitions & Share \\
\midrule
5--19          &  7 &  6.7\% \\
20--49         & 22 & 21.0\% \\
50--99         & 28 & 26.7\% \\
100--199       & 29 & 27.6\% \\
$\geq 200$     & 19 & 18.1\% \\
\midrule
All             & 105 & 100\% \\
\bottomrule
\end{tabular}
\end{table}

\subsection{Corpus Composition and Difficulty}
\label{app:corpus}

Section~\ref{sec:setup} summarizes the corpus along repository-domain and observed difficulty axes. Here we describe the corresponding grouping criteria and provide representative transitions. The released materials include the full per-transition assignments and underlying model scores.

\paragraph{Difficulty stratification.} For each transition, we compute avg@3 $F_1$ for each of the six maintenance models and average the six values to obtain an observed tractability score. We classify transitions as \textsc{Hard} ($<0.40$), \textsc{Medium} ($[0.40,0.65)$), and \textsc{Easy} ($\geq0.65$), yielding 25, 60, and 20 transitions, respectively. The thresholds fall in relatively sparse regions of the observed score distribution and are used only for descriptive analysis. Mean tractability scores are 29.9\% for \textsc{Hard}, 53.0\% for \textsc{Medium}, and 74.5\% for \textsc{Easy}. Neither release-patch size nor $|G_{\Delta}|$ enters the definition.

Table~\ref{tab:task-difficulty} lists representative transitions from the two ends of this observed difficulty spectrum.

\begin{table}[H]
\centering
\caption{Representative hardest and easiest transitions ranked by mean avg@3 $F_1$ across the six maintenance models. Gold lines reports $|G_{\Delta}|$.}
\label{tab:task-difficulty}

\footnotesize
\setlength{\tabcolsep}{4pt}
\begin{tabular}{llrr}
\toprule
Group & Transition & Gold lines & Mean $F_1$ \\
\midrule
Hard & \texttt{DeepSpeed-0.17.6-0.18.0} & 9   & 5.9  \\
Hard & \texttt{peft-v0.15.0-v0.16.0} & 29   & 11.5 \\
Hard & \texttt{agentmemory-0.9.13-0.9.16} & 5 & 12.1 \\
Hard & \texttt{pydantic-v2.7.1-v2.9.1} & 29 & 18.7 \\
Hard & \texttt{datasets-3.0.0-4.0.0} & 144   & 21.6 \\
\midrule
Easy & \texttt{openhuman-v0.53.35-v0.53.40} & 35 & 88.9 \\
Easy & \texttt{langchain-langchain-core==1.0.7-langchain-core==1.1.0}
     & 97 & 85.3 \\
Easy & \texttt{milvus-v2.5.10-v2.5.15} & 72 & 83.1 \\
Easy & \texttt{oh-my-pi-15.1.3-15.2.4} & 60 & 79.9 \\
Easy & \texttt{warp-v1.8.0-v1.8.1} & 199 & 76.8 \\
\bottomrule
\end{tabular}
\end{table}

\paragraph{Domain composition.} Each of the 57 repositories is assigned to one of five domains, and every transition inherits the domain of its repository. The corpus contains 19 agent and RAG framework repositories with 33 transitions, 15 training and inference infrastructure repositories with 32 transitions, 9 data and classic ML repositories with 17 transitions, 9 compiler, kernel, and systems repositories with 16 transitions, and 5 general-purpose library and application repositories with 7 transitions.

\paragraph{Domain-level outcomes.} Mean $F_1$ varies across these domains: 53.8\% for agent and RAG frameworks, 46.4\% for training and inference infrastructure, 57.8\% for data and classic ML, 51.6\% for compiler, kernel, and systems software, and 49.5\% for general-purpose libraries and applications. Training and inference infrastructure is disproportionately represented in the \textsc{Hard} stratum, contributing 9 of the 25 \textsc{Hard} transitions but only 2 of the 20 \textsc{Easy} transitions. These domain assignments are used only for descriptive corpus analysis and do not enter the construction of $G_{\Delta}$, the maintenance runs, or any headline metric.

\end{document}